\pdfoutput=1
\documentclass[11pt]{article}

\usepackage{acl}
\usepackage{booktabs} % For professional quality tables
\usepackage{caption} % For customizing captions
\usepackage{subcaption}
\usepackage{placeins}
\usepackage{times}
\usepackage{latexsym}
\usepackage{graphicx}
\usepackage{booktabs}
\usepackage{amsmath}
\usepackage{multicol}

\usepackage[T1]{fontenc}
\usepackage[utf8]{inputenc}

\usepackage{microtype}

\usepackage{inconsolata}

\title{Noise Correction on Subjective Datasets}

\author{Uthman Jinadu \\
Department of Computer Science, \\
  Georgia State University \\
  Atlanta GA 30303 \\
  \texttt{ujinadu1@gsu.edu} \\\And
  Yi Ding \\
  Department of Computer Science, \\
  Georgia State University \\
  Atlanta GA 30303 \\
  \texttt{yiding@gsu.edu} \\}

\begin{document}
\maketitle
\begin{abstract}
% Accounting for the opinions of all annotators of a dataset is critical for fairness. However, when annotating large datasets, individual annotators will frequently provide thousands of ratings which can lead to fatigue. Additionally, these annotation processes can occur over multiple days which can lead to an inaccurate representation of an annotator's opinion over time. 
Incorporating every annotator's perspective is crucial for unbiased data modeling. Annotator fatigue and changing opinions over time can distort dataset annotations. To combat this, we propose to learn a more accurate representation of diverse opinions by utilizing multitask learning in conjunction with loss-based label correction. We show that using our novel formulation, we can cleanly separate agreeing and disagreeing annotations. Furthermore, this method provides a controllable way to encourage or discourage disagreement. We demonstrate that this modification can improve prediction performance in a single or multi-annotator setting. Lastly, we show that this method remains robust to additional label noise that is applied to subjective data. 
\end{abstract}

\section{Introduction} 
The practice of enhancing label accuracy in datasets through the use of multiple annotations per sample is found in works such as \citet{snow2008cheap}. This method capitalizes on the collective expertise of annotators to improve data quality. The reason is due to the assumption that individual annotators make mistakes and a collection of annotators can provide better-quality labels. Traditional methods for reconciling differences are to use techniques such as majority voting, averaging, or expert opinions \cite{waseem2016hateful}. These methods are designed to arrive at a single `ground truth' for training supervised learning models. However, in tasks of a subjective nature where a single `correct' answer may not exist, these practices can compromise the diverse perspectives of individual annotators. These issues are found across many domains such as in medical \cite{cheplygina2018crowd}, social \cite{ding2022impact}, and others \cite{uma2021learning}.

To improve the fair representation of annotators' opinions, datasets have been built to include individual annotations. These datasets contain labels that are not aggregated through majority voting. This enables the capturing of opinions of annotators who would otherwise have been removed due to the aggregation function. Multitask learning is one way to model this data as demonstrated by \citet{davani2022dealing}, where each task is to predict an annotator's individual label. However, these methods do not account for possible mistakes that individual annotators might make during the annotation process. Datasets such as the GabHateCorpus by \citet{kennedy2018gab} and GoEmotions by \citet{demszky2020goemotions} can contain thousands of annotations by a single annotator. Mistakes such as misidentifying emotions due to annotation exhaustion would not be surprising. Furthermore, annotation of a large number of samples may occur over multiple days and introduce temporal distribution shifts of opinion. 

We propose to address this problem by introducing loss-based label correction into a multitask learning setting. A fundamental property of loss-based label correction works by exploiting the network memorization effect by \citet{arpit2017closer}. They found that networks tended to learn simple patterns before learning more complex ones. This has been used extensively within the noisy learning community such as in \citet{arazo2019unsupervised}. These methods first utilize the memorization effect to detect mislabeled instances. Then they correct these possible sources of mislabeling using techniques such as incorporation of a network's own guess. 

Motivated by this, we propose a novel formulation of multitask learning with label correction. However, a main challenge we try to overcome is that naive applications of these methods can erase the diverse perspectives of annotators. This is because the original technique makes use of a sample's loss to determine whether it is correctly or incorrectly labeled. On a dataset with subjective labels, we find that higher loss samples are associated with minority opinions which complicates the noisy sample detection process. We propose a novel method to address this by strengthening or weakening a model's guess of the true label. We find that our method is robust to added noise while also maintaining diverse opinions. We show that this is useful when modeling subjective datasets with differing agreement properties. We highlight our contributions as follows:
\begin{itemize}
    \item We present a novel formulation of multitask learning with loss-based noise correction.
    \item We demonstrate that we can separate agreeing and disagreeing annotations to detect noise and disagreement. 
    \item We demonstrate that this modification can improve prediction performance in a single or multi-annotator setting. Additionally, this method remains robust to additional label noise that is applied to subjective data.
    \item We introduce a hyperparameter to control the degree of label correction due to variability in label properties for different annotation tasks. We show that this has a noticeable effect on performance.
\end{itemize}

\section{Background}

\subsection{Multitask Learning for Disagreement}
Disagreement among annotators, a common occurrence in the annotation process, signifies differing opinions or interpretations when labeling a specific task. Addressing annotator disagreement is a fundamental part of enhancing machine learning model accuracy. Traditionally, reliability has been used as a measure of dataset quality, however, this measure emphasizes having a single ground truth. This is problematic for capturing diverse opinions. To address these problems, researchers have studied various underlying sources of such disagreements, focusing on two primary types: random variation and systematic disagreement \cite{krippendorff2011agreement, dumitrache2015crowdsourcing, aroyo2013crowd}.

In many real-life scenarios, obtaining the true label for training—considered the gold standard or objective ground truth—is impractical, too costly, or tedious. We often rely on labels from multiple sources, which may introduce noise and significant variation. This variation underlines the challenge of applying supervised learning algorithms without a definitive gold standard, especially when label sources vary in reliability and accuracy \cite{raykar2010learning}.

Random variation encompasses the inherent unpredictability in human judgments, which can introduce randomness into machine learning algorithms, impacting their outcomes \cite{fernandes2023bridging}. This type of disagreement might stem from individual biases or the subjective nature of a task such as due to demographic properties \cite{ding2022impact}.

On the other hand, systematic disagreement is more nuanced, arising from consistent inaccuracies or biases in the machine learning models themselves. Systematic disagreement can also arise from multiple sources or due to multiple reasons. As demonstrated by Krippendorff, this can result from systematic factors that skew the models, leading to persistent errors in interpretation \cite{krippendorff2008systematic}.

Understanding both types of disagreement is crucial for developing robust machine learning models, as it improves the reliability of annotations. Classifier performance and resilience are influenced by the level of annotators' agreement, which is important when selecting training and test data \cite{leonardelli2021agreeing}. Ongoing research continues to enhance methods for identifying and modeling disagreements, thereby improving the quality and representativeness of annotated data.

\subsection{Annotator Noise}
Noise correction refers to the systematic process of identifying and rectifying errors or inconsistencies within a dataset, often termed error correction. Its primary objective is to enhance both the quality and reliability of the data by either eliminating or minimizing the influence of noise \cite{Zhan2019}.

This noise may include random errors, artifacts, or inaccuracies that can significantly distort the underlying patterns and relationships within the data. By employing various correction techniques tailored to the specific type and source of the noise, researchers can obtain a cleaner, more accurate representation of the information.

Traditional categorical cross-entropy loss is not well-suited for this task as it tends to fit the noise rather than the underlying data distribution \cite{zhang2016understanding}. The hard bootstrapping loss introduced by \citet{reed2015bootstrapping} offers a solution by augmenting the standard cross-entropy loss with a perceptual term. This addition helps adjust the training objective, making it more robust to label noise.

Deep neural networks demonstrate a two-phase learning process when trained on datasets with noisy labels. In the "early learning" phase, they tend to pick up correct and general patterns. As training continues, they start to memorize incorrectly labeled examples. This behavior reflects their initial ability to capture generalizable data patterns before gradually fitting to noise, which challenges their generalization capability on new data. This property has been used by many works to improve representation learning under noisy conditions \cite{arazo2019unsupervised, liu2020early, li2020dividemix, nishi2021augmentation}.

Effective noise correction is essential in data preprocessing, especially in machine learning, where data quality significantly impacts model performance and interpretability. This ongoing challenge involves carefully eliminating genuine noise without losing critical variations or introducing new biases \cite{arazo2019unsupervised}.

\subsection{Uncertainty Quantification}
Machine learning models' uncertainty is particularly useful in subjective areas, helping to determine when human oversight is needed, such as in content moderation, where uncertain predictions trigger human review to maintain content standards. \cite{ghandeharioun2019characterizing} \cite{chandrasekharan2019crossmod}

Estimating uncertainty in machine learning is crucial but challenging. The simplest method uses Softmax probabilities \cite{hendrycks2017baseline}, but it can yield overconfident predictions for novel inputs, unlike the training data.

Gal's Monte Carlo dropout technique, highlighted in \citet{gal2016dropout}, improves uncertainty estimation by applying dropout not only during training but also during testing. This method calculates output variances to offer a more accurate assessment of prediction uncertainty than single-label probabilities.

\citet{klas2018uncertainty} argues that relying solely on single label probabilities overlooks many elements influencing predictive uncertainty. Instead, employing a range of predicted annotations provides a more refined and thorough assessment of uncertainty in machine learning models.

\section{Methods}
We are given dataset denoted as \( D \), comprising text samples \( X \), and an annotation matrix \( Y \in \mathcal{R}^{2\times A} \). Here, \( X \) represents individual text entries, \( A \) signifies the number of annotators, and \( Y \) is a matrix where each cell \( y_{i} \) stores one hot value of a label class. Due to the non-uniform distribution of annotator input across text instances, missing values in \( Y \) are prevalent. Our goal is to deduce a consensus label for each text instance. 

\subsection{Baseline}

A standard way for optimizing a model for multi-class classification is to utilize the cross entropy loss function. For an input sample $x$, the cross entropy loss is given by:
\begin{equation} \label{eqn:ce}
L_{CE} = - \sum_{m=1}^M y_m \cdot log(p_m(x, \theta))
\end{equation}
where m represents the number of classes and the $y_i$ represents the truth for the ith class. $\theta$ signifies some model.

\subsection{Multi-task learning to account for different opinions}
Individual annotators' opinions can be incorporated into learning the model via multi-task learning. Suppose there are $A$ annotators, we can create $A$ different task heads (fully connected layers) to predict each of the individual annotations. This changes the cross-entropy loss to a summation of the individual outputs for each task head:
\begin{equation}  \label{eqn:mt}
L_{MT} = - \frac{1}{A}\sum_{a=1}^A \sum_{m=1}^M y_{a, m}\cdot log(p_{a,m}(x, \theta)).
\end{equation}
Here, the ground truth values for each sample form a $A \times M$ matrix $\mathbf{y}$. The model will also output a prediction distribution for each annotator. Note that MT loss reduces to the CE loss when the number of annotators is 1.

\subsection{Noise Correction via Loss Modeling}
\citet{arazo2019unsupervised} found that a sample's loss can be used to separate correctly and incorrectly annotated samples. That is, when learning with imperfect data, lower losses corresponded to the correct samples, and higher losses correlated with incorrect annotations. This leads to a two-step algorithm. First, the loss distribution was modeled using a two-component mixture model to generate a weight $w$. The weight indicates how likely a particular sample, based on its loss, is likely to be mislabeled. The intuition is that a model's guess should be integrated for possibly mislabeled data. This gives us a loss function which scaled by this weight:
\begin{align*}
L_{LC} &= -((1 - w) \sum_{m=1}^{M} y_i \cdot log(p_m(x, \theta)) + \\ \notag
& w  \sum_{m=1}^{M} z \cdot log(p_m(x, \theta)) )
\end{align*}

The new label is re-parameterized as a weighted sum of a network's guesses with its ground truth. Here, $z = p_m(x, \theta)$, and represents a network's prediction based on a single round of evaluation and is detached during a training cycle of the network.

\subsection{Multi-task learning with loss-based noise correction}
\label{sec:novel-loss}

\begin{figure}[!ht]
    \centering
     \begin{subfigure}[t]{.9\linewidth}
        \centering 
         \includegraphics[width=.9\linewidth]
         {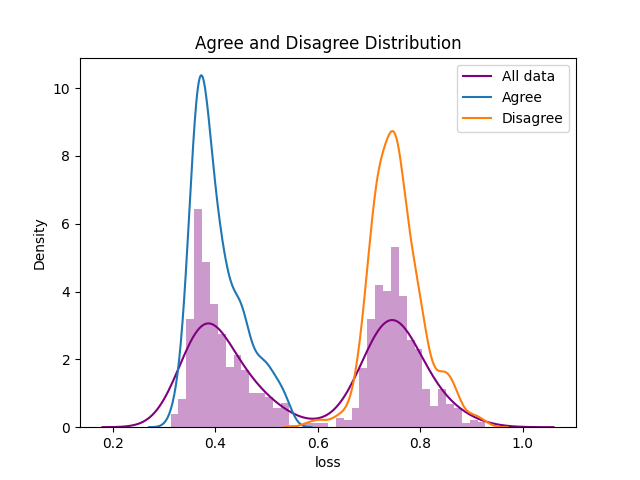}
         \caption{GabHateCorpus loss distribution for a single annotator.}
         \label{fig:ghc_dist}
     \end{subfigure}
     \begin{subfigure}[t]{.9\linewidth}
        \centering
         \includegraphics[width=.9\linewidth]{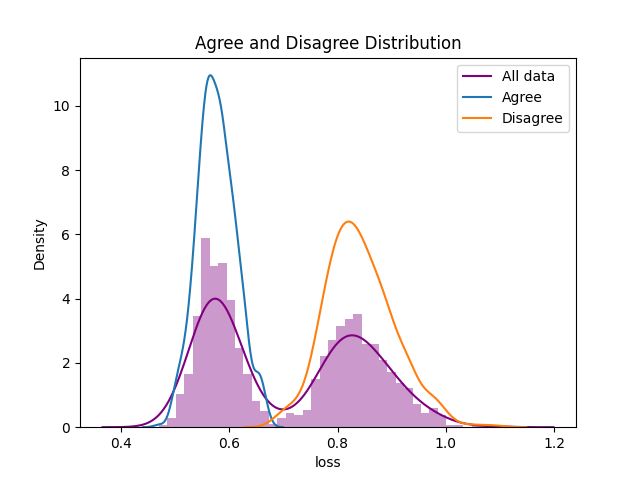}
         \caption{GoEmotions loss distribution for a single annotator.}
         \label{fig:goemo_dist}
     \end{subfigure}
     \caption{Modeling the loss distribution for annotators in multitask learning for a given epoch of training. \citet{arpit2017closer} demonstrated that networks tended to learn correct samples first before memorizing the incorrect samples. In our problem, the semantics of this changes to agreeing versus disagreeing samples. As can be seen, there is a clear separation of samples based on loss.   
     For the dataset depicted in Figure \ref{fig:ghc_dist}, the average proportion of samples in the "Agree" category is 54.57\%, while the "Disagree" category averages 45.43\%. Within the "Agree" distribution, 55.43\% of the samples conform to the baseline majority. In contrast, within the "Disagree" distribution, 48.94\% of the samples align with the majority view. 
     For the dataset illustrated in Figure \ref{fig:goemo_dist}, the "Agree" category comprises an average of 51.69\% of the samples, whereas the "Disagree" category averages 48.31\%. Within the "Agree" distribution of the GoEmotions sample, 52.84\% of the samples align with the baseline majority. Conversely, in the "Disagree" distribution, 48.65\% represent the majority view.
     Additionally, this demonstrates that our method can preserve diverse opinions. This also conforms with the previous understanding that networks tend to be over-confident with predictions since the lower losses are more aligned than higher losses with the majority case.}
     \label{fig:loss_dist}
\end{figure}

We can now combine equations $L_{MT}$ and $L_{LC}$ from above to create a multi-task label correction loss:
\begin{small}
\begin{align*} 
L_{MLC} &= \frac{1}{A}\sum_{a}^A - ( (1 - w) \sum_{m=1}^{M} y_i \cdot log(p_{a,m}(x, \theta)) + \\ \notag
&  w  \sum_{m=1}^{M} z_{a,m} \cdot log(p_{a,m}(x, \theta)) )  \\ \notag
L_{MLC} &= - \frac{1}{A}\sum_{a=1}^A \sum_{m=1}^M (1 - w) y_{a, m}\cdot log(p_{a,m}(x, \theta)) - \\ \notag 
& \quad \frac{1}{A}\sum_{a=1}^A \sum_{m=1}^M w z_{a, m} \cdot log(p_{a,m}(x, \theta)) \\ \notag
L_{MLC} &= (1 - w) * (L_{MT}) + (\psi * w  * L_{G}) 
\end{align*}
\end{small}

Additionally, we propose to introduce a subjectivity parameter $\psi$ parameter to tune the impact of a network's self-guess loss term $\psi \cdot L_G$. 
Intuitively, as shown in Figure \ref{fig:loss_dist}, higher values would push more disagreeing labels to the majority class. This can be tuned as a way to balance agreement with disagreement and noise based on dataset properties.

\subsection{Manifold Mixup}
Integrating the manifold mixup technique into our baseline and multitask loss correction scenarios, as delineated by \citet{verma2018manifold}, involved strategically selecting a random layer from the BERT model. This choice facilitated the application of mixup at the embedding level, addressing the challenges of working with textual data which inherently limits direct input interpolations. Manifold mixup alters the representations linked to layers preceding and succeeding the mixing procedures. 

\section{Experimental Setup}
We evaluate each of the four different optimization methods above in these experiments. 

\begin{itemize}
    \item \textbf{Baseline} corresponds to a network with a single output head optimized to the majority label using $L_{CE}$. 
    \item \textbf{Baseline + Loss Correction} injects our loss correction technique onto the majority label optimization method ($L_{LC})$. The manifold mixup method was adopted.
    \item \textbf{Multitask} involves adding $A$ fully-connected layers to the last layer of the network. Each output is trying to predict the annotations of a unique annotator. Since the datasets we are using are missing annotators, these loss terms are zero-ed out and not back-propagated. This corresponds to the ($L_{MT}$) loss.
    \item \textbf{Multitask + Loss Correction} is our proposed method for integrating loss correction ($L_{MLC}$) into a multitask setting. Just like the baseline loss correction scenario, we applied the manifold mixup to this as well.

\end{itemize}

\subsection{Datasets}
In our experiments, we made use of GoEmotions and GabHateCorpus datasets to annotate our subjective tasks.

\textbf{GoEmotions} \cite{demszky2020goemotions} is a large-scale dataset consisting of 58K Reddit comments annotated with 27 different emotion categories, including neutral labels. The dataset is annotated by 82 crowdsourced workers and label noise is minimized by removing labels that have fewer than one annotator. The dataset is designed for multi-class and multi-label emotion classification, however, 83\% of examples have a single label. The dataset is pre-split into a training set with 43,410 entries, a validation set containing 5,426 entries, and a testing set comprised of 5,427 entries. We evaluate the six basic Ekman emotion labels in this work.

\textbf{GabHateCorpus (GHC):} The Gab Hate Corpus (GHC) hate speech dataset developed by \citet{kennedy2018gab}, and consists of 27,665 text entries collected from Gab.com. This public dataset was annotated by 18 specialized annotators, with each entry evaluated by at least 3 different experts. This rigorous process ensures a high level of accuracy and consistency within the GHC. This property is unique due to its emphasis on label quality when given label diversity.

\begin{table*}[t]
\centering
\begin{scriptsize}
\begin{tabular}{p{5cm}ccccc}
\toprule
Text & Annotator Rating & Majority & Agree With Majority? & Multitask \\
\midrule
Yeah, OK then, get your nurse to give you some more of those nice tablets that make you sleep. & Not Hate & Not Hate & Yes  & Not Hate(100\%) \\
\midrule
This creature was predicated his actions were predicted etc etc etc now the media are outraged as predicted.... this is proof that the level of evil we are dealing with is not a person but a large and well financed group & Not Hate & Not Hate & Yes & Not Hate(100\%) \\
\midrule
Farcebook? Should have tanked a long time ago. & Not Hate & Not Hate & Yes  & Not Hate(67.67\%)\\
\midrule
IDK, my favorite was "Icelandic Women are Whores" Happy New Year and Hail Victory & Not Hate & Not Hate & Yes  & Not Hate(100\%) \\
\midrule
Christian cake baker who won SCOTUS case is now being harassed by demented Satanist homosexuals, backed by the state of Colorado. & Not Hate & Hate & No  & Hate(50\%) \\
\bottomrule
\end{tabular}
% \vspace{.5cm}
% \vspace{-3pt}
% \captionsetup{font=footnotesize}
\caption{Samples from the GabHateCorpus Dataset which shows the text, the annotation on the text given a specific annotator, agreement with the majority based on the Agree and Disagree distribution in Figure \ref{fig:loss_dist}, the baseline majority with the percentage of Hate in the annotation. The samples show instances where the baseline differs removes minority opinions that multitask captures.}
\end{scriptsize}
\label{tab:sample_texts_GHC}
\end{table*}

% \vspace{50pt} % Adjust this value as needed

\begin{table*}[t]
\centering
\begin{scriptsize}
\begin{tabular}{p{5cm}ccccc}
\toprule
Text & Emotion & Annotator Rating & Majority & Agree With Majority?  & Multitask\\
\midrule
But also: fuck the daily mail. & Anger & Anger & Anger & Yes   & Anger(100\%) \\
\midrule
You blew it. They played you like a fiddle & Disgust & Not Disgust & Not Disgust & Yes   & Not Disgust(80\%)\\
\midrule
The sin was refusing to impregnate his brother & Disgust & Not Disgust & Disgust & No  & Disgust(66.67\%) \\
\midrule
Ok now that was epic! & Fear & Not Fear & Not Fear & Yes   & Not Fear(100\%) \\
\midrule
I can see why! I would totally be offended too! & Sadness &  Sadness & Not Sadness & No   & Not Sadness(66.67\%)  \\
\midrule
I’ve been abused enough by the Knicks today. & Sadness & Sadness & Sadness & Yes   & Sadness(100\%) \\
\midrule
I just can't believe she had the nerve to call. & Surprise & Surprise & Surprise & Yes   & Surprise(100\%) \\
\bottomrule
\end{tabular}
% \vspace{.5cm}
% \vspace{-3pt}
% \captionsetup{font=footnotesize}
\caption{Samples from the GoEmotion Dataset which shows the text, the annotation on the text by a specific annotator compared to the baseline majority with the percentage of the emotion, and if it agrees with the majority based on the Agree and Disagree distribution in Figure \ref{fig:loss_dist}.
The samples show instances where the baseline distribution is different than the multitask setting.}
\end{scriptsize}
\label{tab:sample_texts_GoEMotions}
\end{table*}
\subsection{Implementation Details}

The classification models were built using the HuggingFace transformers library (version 4.38) \cite{wolf2019huggingface}. Our experimental setup closely resembled that of \citet{davani2022dealing}. For the GabHateCorpus experiments, we trained the models for five epochs using a learning rate of 1e-7. As for the GoEmotions dataset, we utilized a learning rate of 5e-6 and trained the models for five epochs. We used this to obtain the baseline and multitask for each of the datasets and then we introduced the label correction similar to how \citet{arazo2019unsupervised} was implemented. We used the pre-trained \textit{Bert base-uncased} model as the underlying architecture. As we do not have access to the original multi-task code some results are different than \citet{davani2022dealing}. Optimization was conducted using Stochastic Gradient Descent (SGD) with a momentum of \(0.9\), weight decay of \(0.01\), and a batch size of \(64\).  

Additionally, we integrate some components that are necessary to train for loss correction. A warm-up period using an entropy-based penalty on the confidence term has been found useful to prevent noise overfitting during warm-up periods \cite{pereyra2017regularizing}. We designated a warm-up period of \(2\) epochs for the GoEmotions and  GabHateCorpus datasets. This was necessary to stabilize the training, mitigate poor initialization, and make it start with a smaller learning rate and gradually increase it. We utilized regularization following \citet{tanaka2018joint} and \citet{arazo2019unsupervised} aiming to deter the allocation of all samples to a singular class. We did not implement the full \citet{arazo2019unsupervised} to minimize possible confounding effects of auxiliary techniques. Unless otherwise specified, $\psi$ is set to $0.5$.

\subsection{Noise Injection}
\label{sec:setup_noise_inj}
The datasets used in the evaluation are for binary classification. Therefore, we simply flip the labels by sampling for a subset of data. In this paper, we used 20\% noise rate for the noise experiments. For samples where the labels are from multiple annotators, e.g. 4 votes for true and 1 vote for false, we swap the labels for the samples where we are injecting noise. This is similar to how labels are injected in popular noisy labels literature. Not all annotators annotated every sample. We ignored these annotators when injecting noise and thus these did not have an impact on the loss.

\section{Results}

Due to the nature of this problem, it is important to consider these results within the context of data collection processes. GoEmotions annotators were crowdworkers who spoke English from India. GHC annotation was a two-step process that involved expert opinions. The annotators of GHC were trained and the annotation process was much more controlled. Furthermore, the annotator population was recruited from a smaller sample and consisted of undergraduate students. These processes would cause GHC to be less variable, due to likely similar backgrounds of annotators. On the other hand, using untrained crowdworkers from a foreign country would likely lead to more variability.

With this in mind, we first discuss the comparison of performance metrics in evaluated methods. We then examine the effect of injecting additional noise. Lastly, we examine the effect of adjusting the subjectivity parameter and its effect on results.

\subsection{Comparison of techniques}

\begin{table*}[t]
\footnotesize
\centering
\begin{tabular}{lllllll}
\toprule
& \multicolumn{3}{c}{No Noise} & \multicolumn{3}{c}{20\% Noise} \\
\cmidrule(r){2-4} \cmidrule(l){5-7}
Method & Precision & Recall & F1 & Precision & Recall & F1 \\
\midrule
Baseline & 46.09 $\pm$ 2.9 & 49.96 $\pm$ 5.7 & 47.95 $\pm$ 5.4 & 49.87 $\pm$ 5.6 & 49.78 $\pm$ 5.6 & 41.67 $\pm$ 4.7 \\
Baseline + Label Correction & 50.11$\pm$3.2 & 51.32$\pm$3.3 & 49.34$\pm$5.6 & 50.50$\pm$3.2 & 50.12$\pm$3.2 & 45.97 $\pm$5.2 \\ \midrule
Multitask & 49.56$\pm$5.6 & 49.86 $\pm$ 5.0 & 45.57$\pm$2.9 & 47.70$\pm$3.0 & 47.95$\pm$3.1 & 45.62$\pm$2.9 \\
Multitask + Label Correction & \textbf{51.54} $\pm$3.3 & \textbf{51.71}$\pm$2.8 & \textbf{50.3} $\pm$3.2 & \textbf{52.32} $\pm$3.3 & \textbf{51.12}$\pm$2.5 & \textbf{51.55}$\pm$2.3 \\
% Dual Network Co-refinement(Baseline) & 46.56 & 48.43 & 52.36 & 48.65 & 48.49 & 55.45 \\
\bottomrule
\end{tabular}
% \vspace{.5cm}
% \vspace{-4pt}
% \captionsetup{font=footnotesize}
\caption{Performance comparison on the GabHateCorpus dataset when compared to majority based ground-truth. Our proposed method of incorporating label correction into multitask learning had the highest score. Our method is also robust to label noise under conditions of label subjectivity. This experiment is run with 5 different seeds and we report the averages.}
\label{tab:ghc_results}
\end{table*}
% \vspace{-.2cm}

\begin{table}[t]
\scriptsize % Setting font size for both tables globally
% \begin{minipage}[t]{0.48\linewidth} % Adjusted for alignment and given space for separation
\centering
\begin{tabular}{l|ll|ll}
\toprule
\multicolumn{5}{c}{No Noise} \\
\cmidrule(r){1-5}
Emotion & Baseline & Baseline + LC & Multitask & Multitask + LC \\
\midrule
Anger & 42.21 $\pm$ 3.7 & 43.35$\pm$3.6 & 56.31$\pm$3.6 & \textbf{66.68}$\pm$4.6  \\
Disgust & 32.4$\pm$6.1 & 34.18$\pm$3.1 & 58.16$\pm$2.4 &  \textbf{64.65}$\pm$0.5  \\
Fear & 44.45$\pm$6.0 & 49.09$\pm$5.7 & 58.96$\pm$5.5 & \textbf{69.80}$\pm$6.1  \\
Joy & 52.17$\pm$5.3 & 54.63$\pm$5.0 & 60.91$\pm$5.2 & \textbf{66.97}$\pm$4.4 \\
Sadness & 50.29$\pm$5.7 & 51.84$\pm$3.8 & 62.16$\pm$5.8 & \textbf{67.86}$\pm$3.9  \\
Surprise & 44.32$\pm$5.1 & 46.86$\pm$6.1 & 61.63$\pm$5.0 & \textbf{68.93}$\pm$3.5 \\
\bottomrule
\end{tabular}
% \captionsetup{font=scriptsize}
\caption{Performance comparison on the GoEmotions dataset. Binary classification accuracy is reported for each emotion. For each emotion considered, we demonstrate improved performance when $\psi$ is fixed to 0.5.}
\vspace{.2cm}
\label{tab:goemo_results}
% \end{minipage}%
\hfill % Adds horizontal space between minipages if needed and helps in alignment
% \begin{minipage}[t]{0.48\linewidth} % Adjusted for alignment
\centering
\begin{tabular}{l|ll|ll}
\toprule
\multicolumn{5}{c}{20\% Noise} \\
\cmidrule(r){1-5}
Emotion & Baseline & Baseline + LC & Multitask & Multitask + LC\\
\midrule
Anger & 46.58$\pm$5.9 & 49.93$\pm$3.0 & 44.28$\pm$5.0 & \textbf{48.26}$\pm$3.0  \\
Disgust & 44.3$\pm$3.3 & 48.94$\pm$4.9 & 43.89$\pm$1.1 & \textbf{46.27}$\pm$4.9   \\
Fear & 44.38$\pm$3.6 & 49.08$\pm$3.5 & 44.11$\pm$4.8 & \textbf{47.05}$\pm$3.1  \\
Joy & 46.61$\pm$4.9 & 50.35$\pm$6.1 & 43.62$\pm$5.2 & \textbf{48.83}$\pm$3.6 \\
Sadness & 45.51$\pm$3.6 & 49.47$\pm$4.9 & 43.58$\pm$3.8 & \textbf{47.77}$\pm$1.9  \\
Surprise & 44.53$\pm$1.2 & 49.24$\pm$2.9 & 43.83$\pm$3.6 & \textbf{49.01}$\pm$3.4 \\
\bottomrule
\end{tabular}
% \captionsetup{font=scriptsize}
\caption{Performance comparison on the GoEmotions dataset when injected with 20\% label noise. Our method shows robustness to noise injection as well. The observed performance reduction in multitask and multitask+LC models, compared to the baseline and baseline+LC, upon noise injection might be attributed to the specific method of integrating noise, the number of annotators of the GoEmotion dataset, and other inherent characteristics of the GoEmotion dataset. This suggests exploring alternative noise simulation techniques might mitigate the impact on model performance, particularly in datasets with complex attributes. }
\label{tab:goemo_noise_results}
% \end{minipage}
\end{table}
% \vspace{-1cm}

We present the GabHateCorpus results in Table \ref{tab:ghc_results} and the results of GoEmotions in Table \ref{tab:goemo_results}. All metrics reported are compared to the majority label on the ground truth. As can be seen in the GHC results, our method demonstrates improvements across all metrics when compared against the baselines. Additionally, we see that baseline methods based on the majority also perform worse than baseline + label correction. These results indicate that regardless of the number of annotators, it may be beneficial to incorporate label correction. 

For GoEmotions results, we see that for Multitask+LC, all six cases exceed the multitask results. This indicates that the loss correction is having some effect. Most notably, there is a significant improvement in performance between the majority and annotator (multitask) conditions. We hypothesize that this is due to two factors.  One plausible explanation for this phenomenon may be the inconsistency in the number of annotators who labeled each text instance in the dataset. This pattern was also noted by \citet{davani2022dealing}. Such variance could introduce an additional layer of complexity or uncertainty into the model, affecting its ability to efficiently generalize from the data. This opens up avenues for future research to understand how the number of annotators, and the possible divergence in their annotations, influences the effectiveness of loss correction methods in multi-task learning setups. While we observed an improvement in the multitask scenario, we believe the application of manifold mixup is also contributory to this, our multitask results exceed the results presented in the previous work. Secondly, additional investigation is warranted for the effect of the initial model which further confounds the results.

\subsection{Classification of Subjective Labels in the Presence of Noise}

We compare the effect of noise for each classification task on GHC as well as GoEmotions. In this paper, we conduct experiments with a 20\% noise rate injected based on the process discussed in \ref{sec:setup_noise_inj}. As can be seen in Table \ref{tab:ghc_results} our method is robust to noise injection even when accounting for multiple annotators. It is also interesting to note that both the majority-based methods for aggregating labels (Baseline Majority and Baseline + Label Correction), showed a large drop in performance. This highlights the importance of training multiple annotators as a way to mitigate the effect of noise.

We see that for the results of GoEmotions in Table \ref{tab:goemo_noise_results}, there is a significant drop in the majority label annotations. However, the multitask cases all showcase a smaller drop in performance than the majority label techniques. This highlights the need to account for multiple annotator opinions.

\subsection{Effect of Subjectivity Parameter}

\begin{table*}[t]
\centering
\footnotesize
\begin{tabular}{llll|lll}
\toprule
          & \multicolumn{3}{c}{\textbf{No Noise}} & \multicolumn{3}{c}{\textbf{20\% Noise}} \\ \cmidrule(r){2-7}
  Label type& $\psi =$ 1         &  $\psi =$ 0.5        &  $\psi =$ 0.25       &  $\psi =$ 1           &  $\psi =$ 0.5         &  $\psi =$ 0.25        \\ \midrule
GHC       & 47.86       & 50.3      & 48.13      & 45.8       & 51.55       & 48.71       \\ \midrule
Anger     & 55.05       & 66.68      & 56.96     & 45.40       & 48.26       & 45.49       \\
Disgust   & 56.74       & 64.65      & 55.5      & 44.64       & 46.27       & 44.82       \\
Fear      & 57.55       & 69.80      & 57.55      & 44.80       & 47.05       & 43.67       \\
Joy       & 55.57       & 66.97      & 57.83      & 44.45       & 48.83       & 44.53       \\
Sadness   & 53.42       & 67.86      & 53.41      & 44.68       & 47.77       & 44.73       \\
Surprise  & 55.39       & 68.93      & 60.77      & 45.36       & 49.01       & 43.74      \\ \bottomrule
\end{tabular}
% \vspace{.5cm}
% \vspace{-3pt}
% \captionsetup{font=small}
\caption{Comparison of accuracy for GHC and GoEmotions for different parameters of $\psi$. Adjusting $\psi$ has a clear effect on GHC and GoEmotions. Both GHC and GoEmotions appear to perform the best in general for $\psi = 0.5$. We believe the application of the manifold mixup to the training has some effects on this. This indicates a difference in the label patterns for subjectivity and noise.}
\label{tab:subj_param}
\end{table*}

We present the results exploring the subjectivity parameter $\psi$ in Table \ref{tab:subj_param}. A higher $\psi$ will lead to more conforming predictions, while a lower $\psi$ will lead to less conforming predictions. Our investigation into the optimal setting for the subjectivity parameter ($\psi$) across GHC and GoEmotions datasets demonstrates that a $\psi$ value of 0.5 consistently achieves the highest performance (relative to the majority vote) for our experimental setup. This result highlights the importance of a balanced approach to noise correction across diverse datasets, showing that a $\psi$ value of 0.5 effectively manages label disagreements and boosts model accuracy. In addition to the majority-based patterns we see in Table \ref{tab:subj_param} we present annotator-level results in Table \ref{tab:subjectivity-acc-var}. As can be seen, there is a direct correlation between the annotator-level labeling variance and the $\psi$ parameter. This appears to be more visible in the dataset GHC as it is more balanced than GoEmotions. To calculate variance across all annotations, we follow the same formula as previous work: 
\[
\sigma^2(\bar{y_i}) = \frac{\sum[y_{ij}=1]\sum[y_{ij}=0]}{|\bar{y_i}^2}.
\]
The inherent noise in the dataset suggests that some correction of annotation mistakes is necessary, but it will compromise prediction diversity. The results reported are averaged over two runs.

% Please add the following required packages to your document preamble:
% \usepackage[table,xcdraw]{xcolor}
% Beamer presentation requires \usepackage{colortbl} instead of \usepackage[table,xcdraw]{xcolor}
\begin{table}[t]
\centering
\footnotesize
\begin{tabular}{ll|lll}
\toprule
\textbf{Dataset} & \textbf{Metric}     & \textbf{$\psi = 0.25$}       & \textbf{$\psi = 0.5$}        & \textbf{$\psi = 1$}                                  \\ \midrule
\textbf{GHC}     & Anno. Acc.  & \multicolumn{1}{r}{59.48} & \multicolumn{1}{r}{63.82} & \multicolumn{1}{r}{58.67} \\
\textbf{GHC}     & Pred. Var. & \multicolumn{1}{r}{14.38} & \multicolumn{1}{r}{13.59} & \multicolumn{1}{r}{13.01}                         \\ \midrule
\textbf{GoEmo}   & Anno. Acc.  &  \multicolumn{1}{r}  {49.92}                & \multicolumn{1}{r}  {50.11}                          & \multicolumn{1}{r}  {49.85}                                                  \\
\textbf{GoEmo}   & Pred. Var. & \multicolumn{1}{r}  {17.74}                          & \multicolumn{1}{r}  {17.53}                           & \multicolumn{1}{r}  {17.46}       \\ \bottomrule                                          
\end{tabular}
\caption{Annotator-level accuracy and variance when adjusting for $\psi$ reported as percentages. Lower $\psi$ encourages diversity of annotation while higher $\psi$ encourages agreement. As can be seen, prediction variance is highest when the $\psi$ value is lowest. The annotator-level accuracy is also the highest when the $\psi$=0.5 and reflects the same patterns seen in majority vote results shown in Figure \ref{fig:variance_psi}. These results demonstrate that we can tune for opinion and noise properties in a controllable way.}
\label{tab:subjectivity-acc-var}
% \vspace{-.5cm}
\end{table}

While this value may change depending on hyperparameters, dataset, architecture, and other properties; it highlights the need to take a balanced approach when adjusting for noise correction on the dataset with subjective annotations. Setting the value too high may lead the network to make incorrect guesses and may lead to increased variance and inaccuracies. Setting it to an `optimal' value may cause the dataset to predict the majority cases more, however, this may not be desirable in all situations and thus our $\psi$ parameter provides a way to control for this property.

%This finding suggests that adjusting $\psi$ to this middle ground is a straightforward yet efficient strategy for enhancing performance, making it a potentially universal solution for dealing with label variance in different contexts.

\subsection{Loss Separation}
We showcase a clear bimodal distribution to illustrate the challenge of separating noise and opinion in Figure \ref{fig:loss_dist}. We see that for a specific annotator, there are peaks for agreeing and disagreeing samples. Note that these are semantically different than the noisy label's definition of correct and incorrect. In this paper, we refer to incorrect annotations as a label that the annotator would not typically ascribe to a sample due to the dataset properties. We assume that incorrect answers are more likely to fall in the distribution of disagreeing labels. We calculate the proportion of labels within each mode to see how many of them agree with the majority annotation and how many disagree. Our proposed technique attempts to control the level of noise correction to account for these added complications.

\subsection{Model Uncertainty}

\begin{figure*}[!t]
     \centering
         \centering
         \includegraphics[width=.75\linewidth]{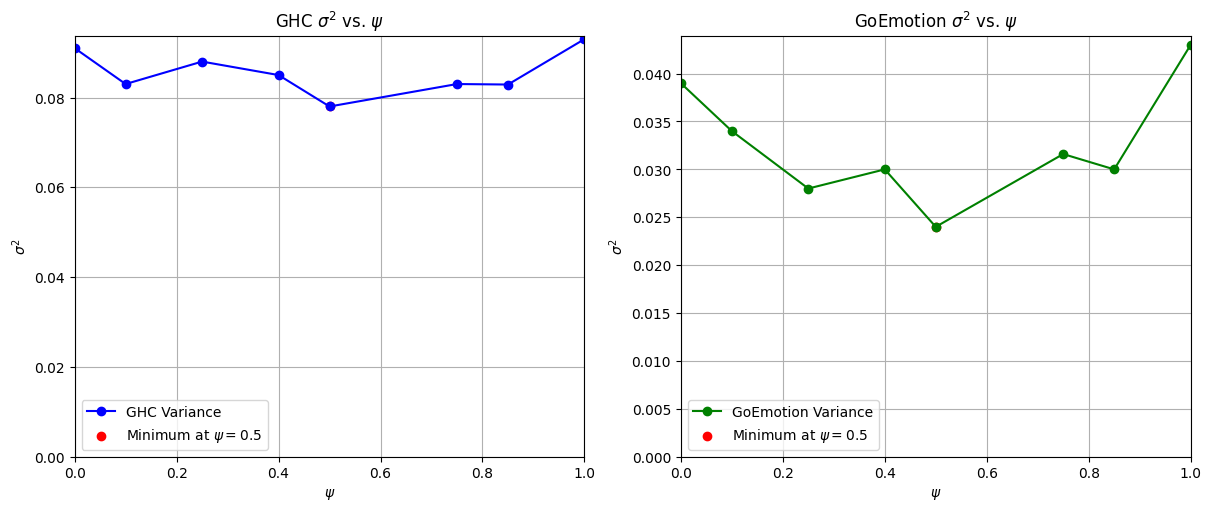} 
    % \captionsetup{font=small}
     \caption{The variance in model predictions against majority labels for the \texttt{GHC Multitask} and \texttt{GoEmotion Multitask} models. Notably, at a $\psi$ value of 0.5, variance reaches its minimum, indicating the highest agreement among annotators. This point reflects the optimal balance in the loss correction strategy, enhancing model alignment with consensus sentiment.}
     \label{fig:variance_psi}
\end{figure*}

In our empirical evaluations, we observed an increase in the agreement between model predictions and majority labels upon applying the loss correction strategy to multitask scenarios. 
The provided graphs in Figure \ref{fig:variance_psi} illustrate the effect of the loss correction method on the variance of predictions for two multitask models across different values of $\psi$. 

For the GHC Multitask model depicted on the left of Figure \ref{fig:variance_psi}, there is a visible trend where the variance decreases to its lowest point at $\psi$ = 0.5. This represents an 18\% increase in the agreement between the model's predictions and the majority labels. The GoEmotion Multitask model shown on the right shows a corresponding improvement of approximately 14\% improvement in prediction consistency. The other points on the graph for GoEmotion vary more significantly, with some psi values leading to higher variances, which suggests greater fluctuation in agreement levels at those points.

Both graphs highlight the optimal setting of $\psi$ to $0.5$ for achieving the most reliable predictions, as indicated by the minimum variance values at this point, suggesting the effectiveness of the loss correction strategy in enhancing model reliability. This process quantifies the consistency of model predictions where a lower variance suggests a higher agreement with majority. This highlights the technique's effectiveness in boosting the models' reliability and applicability, especially in scenarios characterized by subjective judgment and high uncertainty.

\section{Discussion}
Noisy-label tolerant techniques make the assumption that there is a single ground truth. Many of these techniques, e.g. \citet{arazo2019unsupervised}, utilize a label's loss to determine whether a sample is correctly labeled or not. That is higher loss is generally incorrectly labeled. However, when opinion is involved, we cannot naively do this. As we show in our results on Figure \ref{fig:loss_dist}, the semantics of a samples' loss changes to majority or minority opinions. That is, higher loss is associated with minority opinions while lower loss is associated with majority ones.

At a per-annotator level, there are still mistakes being made. Thus we need to find a way to utilize these loss properties when learning representations. To do so, we propose a novel loss function (Section \ref{sec:novel-loss}) which models the predictions for each annotator which factorizes out nicely. We then introduce an additional parameter to control the strength of this noise correction process to balance noise with disagreeing opinions. This methodology is particularly relevant for subjective tasks like online hate speech detection on platforms such as Reddit and Twitter, striving for models that effectively reduce noise and enhanced opinion modeling. Our experimental outcomes underscore the significance of adopting a balanced strategy for noise correction across varied datasets (Table \ref{tab:subjectivity-acc-var}). It also shows cases that annotator-level mistakes exist and can be corrected for (Tables \ref{tab:ghc_results}, \ref{tab:goemo_results}, \ref{tab:goemo_noise_results}). Accounting for these factors leads to an improvement in performance metrics at both annotator and aggregate levels while maintaining prediction diversity.

%In our study, we introduced a method for correcting noise in multitasking scenarios, focusing on predicting individual annotator labels within subjective datasets, specifically utilizing GabHateCorpus and GoEmotion. We differentiated between "agreeing" annotations—those that form the majority and exhibit lower loss—and "disagreeing" annotations, which represent minority opinions with higher loss. Our approach initially employs majority voting as a baseline to consolidate annotator opinions, subsequently advancing to multitask learning for predicting distinct annotations. Noise correction, aimed at identifying and amending dataset inaccuracies, seeks to improve data quality and reliability by minimizing noise. 

\section{Conclusion}
In this study, we developed a method for loss-based label correction for a multitask setting. Specifically, this is applied to subjective datasets from multiple annotators. We show that our approach preserves disagreeing labels and that the performance increases under some conditions. We also demonstrated that it is robust to added noise. Lastly, we demonstrate that adjusting the degree of loss correction is important and has a large impact on performance. This is beneficial for future work in collecting subjective labels at scale.

\section{Limitations}
% Despite the observable significant effect of $\psi$ on performance especially when the value is set to 0.5, pinpointing the optimal value is complex, with no precise selection strategy currently in place. The parameter not only diminished the variance but also enhanced performance reinforcing our findings in the results reported for the GoEmotions, however, interpreting the significance of these improvements is challenging due to the high imbalance in the GoEmotions dataset. Nonetheless, our results still demonstrate an enhancement over previous studies \cite{davani2022dealing}. This calls for additional investigation to establish a methodical approach for $\psi$ determination, which is essential for the method's adaptability and accuracy across varying datasets.  
Although $\psi$ significantly affects performance, particularly at 0.5, identifying the optimal value remains complex without a clear selection strategy. The parameter reduced variance and improved performance, as evidenced in the GoEmotions dataset, but interpreting these gains is difficult due to the GoEmotions dataset imbalance. Yet, our results still show an improvement over previous studies \cite{davani2022dealing}. This underscores the need for further research to develop a systematic approach for determining $\psi$, crucial for the method's adaptability and precision across different datasets.
Additionally, the typical way to analyze the noise patterns of less subjective datasets is to annotate a subset of the test set at scale. This gives insight into the noise profile and can enable the creation of a label noise transition matrix. To fully evaluate the noise patterns of subjective annotations is much more costly. Annotation subjectivity can vary through time, and with demographic properties. These properties make building a noise transition matrix almost impossible due to the number of factors that can influence the underlying label distribution. Thus, understanding noise for subjective annotation tasks remains an exciting challenge we leave for future work.

\section{Ethics Statement}
In data annotation, capturing the full spectrum of annotator perspectives is important for achieving equitable outcomes. Challenges such as annotator fatigue and variability in judgment over time can obscure the true diversity of opinions within large datasets. To address these issues, our method introduces a novel combination of multi-task learning and loss-based label correction. This method effectively separates matching and mismatching annotations while improving prediction accuracy for both individual and multi-annotator settings. Similar to the use of multi-annotator models in affect detection \cite{alm2011subjective}, our method models diverse opinions for a deeper insight into subjective data.
Additionally, our technique adapts to the intrinsic label noise in subjective annotations, enhancing robustness similar to methods in sarcasm detection that consider annotator-specific thresholds \cite{rakov2013sure}. This refined approach captures the complexity of human judgment, promoting fairer data interpretation in digital annotation.
% For instance, similar to how multi-annotator models have been utilized in affect detection to encompass a wide range of human emotions and interpretations \cite{alm2011subjective}, our method effectively models diverse opinions, yielding a richer understanding of subjective data. 
% Additionally, by adapting to the intrinsic label noise present in subjective annotations, our technique ensures robustness, akin to strategies employed in sarcasm detection where annotator-specific sarcasm thresholds are considered \cite{rakov2013sure}. Ultimately, this refined approach to representing diverse annotator insights promotes fair data interpretation, mirroring the complexity of human judgment in digital annotation processes.

\section{Acknowledgements}
Research was sponsored by the Army Research Laboratory and was accomplished under Cooperative Agreement Number W911NF-23-2-0224. The views and conclusions contained in this document are those of the authors and should not be interpreted as representing the official policies, either expressed or implied, of the Army Research Laboratory or the U.S. Government. The U.S. Government is authorized to reproduce and distribute reprints for Government purposes notwithstanding any copyright notation herein. Research was also supported by a gift award from The Home Depot. We thank the reviewers for their constructive criticism and valuable advice, which have improved the quality of this final paper.

% Bibliography entries for the entire Anthology, followed by custom entries
%\bibliography{anthology,custom}
% Custom bibliography entries only
{
\small
% \clearpage  % or \newpage
\bibliography{anthology, custom}
}

\clearpage
\appendix

% \section{Multitask Label Correction Equation}
\label{sec:appendix}
\end{document}